\documentclass[runningheads]{llncs}
\usepackage[T1]{fontenc}
\usepackage[utf8]{inputenc}

\usepackage{graphicx}
\usepackage{xcolor}
\usepackage{amsmath,amssymb,amsfonts}
\usepackage{bm}
\usepackage{booktabs}
\usepackage{multirow}
\usepackage{makecell}
\usepackage{algorithm}
\usepackage{algorithmic}
\usepackage{float}
\usepackage{tcolorbox}
\usepackage{listings}
\usepackage{microtype}
\usepackage[font=small]{caption}
\usepackage[hidelinks]{hyperref}
\usepackage{orcidlink}

\newcommand{\visibleorcid}[1]{\orcidlink{#1}}

\providecommand{\BIBentryALTinterwordspacing}{}
\providecommand{\BIBentrySTDinterwordspacing}{}

\begin{document}
\title{T2T-LA: A Topology-to-Topology LLM Agent for Graph Learning with Neither Feature Access nor Task Knowledge}


\author{Yongyu Wang\,\visibleorcid{0009-0006-0705-752X}}
\authorrunning{Y. Wang}

\institute{
\email{yongyuw@mtu.edu}
}

\maketitle
\renewcommand{\thefootnote}{}

\renewcommand{\thefootnote}{}

\begin{abstract}
  Graph learning aims to convert data into graph representations, which are fundamental to many problems in machine learning for CAD, where circuits, layouts, designs, and optimization states are often modeled as graph-structured objects. Existing graph learning methods usually rely on carefully designed graph construction rules, extensive parameter tuning, and sophisticated mathematical theory; moreover, achieving good performance often requires task-specific graph construction tailored to the downstream objective. In this work, we study whether a large language model (LLM) can reason about graph structure and infer a useful topology without observing the feature matrix, without knowing the downstream task, and without relying on any carefully designed graph construction algorithm or parameter tuning process. To this end, we propose T2T-LA, a Topology-to-Topology LLM Agent that receives no input other than a set of previously failed topologies and the scores assigned to them by a private scorer. The agent is not told what task or algorithm produces the scores, how these topologies are generated, or what the scores mean. Since none of the observed topologies is satisfactory, T2T-LA cannot simply imitate a good example. Instead, it is forced to infer hidden relationships between graph connectivity patterns and the observed scores, a capability that is particularly relevant to CAD scenarios where useful design structures may be difficult to specify manually. Experimental results show that T2T-LA can generate, in one shot, a graph topology that enables the downstream algorithm to produce a sufficiently good solution, suggesting a new LLM-driven direction for topology reasoning and graph representation learning in ML-for-CAD workflows.
\end{abstract}

\section{Introduction}

Graph-based algorithms play an important role in  machine learning for computer-aided design. In many CAD and electronic design automation (EDA) workflows, design objects such as circuits, netlists, layouts, components, and optimization states can naturally be represented as graph-structured data, where nodes describe design entities and edges describe physical, logical, or functional relationships \cite{Ma2020UnderstandingGI,Lu2022OnAP,Yang2022VersatileMG}. By representing data instances as nodes and relationships as edges, a graph topology can reveal structural information that is difficult to capture from independent data points alone \cite{carey2017graph}. Moreover, representing data as graphs enables research findings from graph theory to be applied to downstream learning, optimization, and design analysis problems \cite{chung1997spectral}.

In graph-based algorithms, graph quality is critical to the performance of the overall algorithm \cite{maier2008influence}. This issue is particularly important in ML-for-CAD scenarios, where an inaccurate topology may distort the structural relationships among design components and negatively affect downstream tasks such as clustering, prediction, optimization, or design-space exploration \cite{Aghdaei2022HyperEFSH,Aghdaei2021HyperSFSH}. Therefore, over the past few decades, substantial research effort has been devoted to better transforming data into graphs, and many methods have been proposed. A common strategy is to construct similarity-based graphs, such as $k$-nearest-neighbor graphs, where each data point is connected to nearby samples according to a predefined distance metric \cite{paredes2006practical}. Despite their simplicity, these methods are highly sensitive to design choices, including the neighborhood size, similarity function, and edge-weighting rule \cite{Dong2011EfficientKN}. Such choices are often task-dependent and difficult to determine in advance, and inappropriate settings may cause the resulting graph to either overlook global structures or contain unreliable local connections \cite{premachandran2013consensus}. In CAD-related applications, this sensitivity can be even more problematic because the most useful structural relationships are not always directly observable from raw features or simple distance metrics \cite{Sajadinia2025HyperEF2S,Cheng2025CirSTAGCS}.

To improve graph quality, researchers have developed increasingly sophisticated graph construction methods. Adaptive-neighbor \cite{liu2018learning} and consensus-based \cite{premachandran2013consensus} methods try to build more reliable topologies by refining local neighborhoods or preserving the geometry of the data manifold. However, these improvements often come at the cost of more complicated algorithmic designs, additional hyperparameters, and higher computational overhead \cite{wang2022scalable}. Optimization-based graph learning methods are even more mathematically involved \cite{pasham2023network,bielefeldt2019system}. Methods such as graphical Lasso, graph-signal-processing-based Laplacian estimation, and spectral-constraint-based graph learning are built upon sophisticated mathematical theories and advanced technical machinery, including sparse inverse covariance estimation, graph signal processing, Laplacian-based optimization, and spectral graph analysis \cite{friedman2008sparse,egilmez2017graph,kalofolias2016learn,ortega2018graph,DBLP:conf/iclr/KalofoliasP19}. Although powerful, these methods are usually computationally expensive, require careful tuning, and may not scale well to large datasets or complex design spaces. For example, many Laplacian estimation methods require at least $O(N^2)$ time per iteration for $N$ data samples \cite{dong2016learning,dong2019learning}, while manifold-based graph construction methods such as those relying on Isomap-style embeddings \cite{tenenbaum2000global} can require $O(N^3)$ time for manifold construction \cite{carey2017graph}. Even methods that reduce the search space using approximate nearest-neighbor graphs still involve nontrivial optimization procedures and may remain inefficient on large-scale data \cite{DBLP:conf/iclr/KalofoliasP19}. Although some recent methods attempt to reduce the computational cost of graphical-Lasso-based topology estimation through spectral perturbation \cite{wang2022scalable}, additional modules, such as customized Algebraic Multigrid (AMG) solvers and eigensolvers, are still needed \cite{Zhao2017SAMGSG,wang2022scalable}.

These limitations motivate a different question: instead of manually designing graph construction rules or solving complex graph-learning optimization problems, can an LLM infer useful topology-level information directly from previously failed graph structures? This question is closely related to ML-for-CAD, where useful design relationships may be hidden, indirect, or difficult to capture using conventional machine learning models \cite{Yang2022VersatileMG}. In practical circuit design and manufacturing processes, many unsatisfactory design structures and failed products are generated \cite{Hu2007PatternSP,Rajaram2004ReducingCS}. However, the information contained in these imperfect or failed designs is rarely fully exploited.

Inspired by the strong reasoning capabilities demonstrated by LLMs across various types of tasks \cite{Kojima2022LargeLM}, we propose T2T-LA, a Topology-to-Topology LLM Agent that studies whether an LLM can infer the hidden relationship between edge connectivity patterns and topology qualities under extremely limited information. Unlike conventional graph learning methods that construct graphs from data features or optimize explicit graph-learning objectives, T2T-LA treats topology generation as a reasoning problem over failed structural examples. Specifically, T2T-LA has no access to node features, does not know the downstream algorithm that uses the graph, is unaware of any graph construction algorithm, and is not provided with the evaluation metric or its meaning. Instead, it only observes a set of previously failed topologies and their corresponding topology scores with unknown meanings, and is asked to infer a new topology that can achieve a high score. Since all observed topologies are unsatisfactory, the agent cannot simply imitate a high-quality graph. Instead, it must compare structural differences among poor topologies and infer which connectivity patterns are more likely to improve the hidden score.

\section{Related Work}

Existing studies have explored the use of LLMs in several graph-related settings. In graph reasoning, LLMs are used to interpret graph structures and solve algorithmic graph problems, such as shortest path, topological sorting, and maximum flow~\cite{wang2023can}. In graph understanding, LLMs are evaluated on whether they can understand graph-structured data and perform tasks such as node classification based on node attributes or structural features, link prediction between nodes, and graph query generation according to user requirements~\cite{guo2023gpt4graph}. In graph structure learning, LLMs are used to denoise noisy connections and refine graph topology~\cite{guo2024graphedit}. In graph contrastive learning, LLMs are used to exploit textual attributes for graph augmentation \cite{fang2024gaugllm}.

In these studies, LLMs have played a satisfactory role in improving graph-based algorithms. For example, by harnessing advanced LLMs for feature-level and structure-level augmentations, the performance of many leading graph contrastive learning methods, such as BGRL \cite{thakoor2021large}, GraphCL \cite{you2020graph}, and GBT \cite{bielak2022graph}, can be significantly enhanced~\cite{fang2024gaugllm}. By strengthening the graph-structure reasoning ability of LLMs through instruction tuning, LLMs can reason about the underlying graph structures, obtain more reliable graph topologies, and improve the accuracy of downstream tasks such as node classification~\cite{guo2024graphedit}.

However, in these studies that apply LLMs to graph-based algorithms, the effectiveness of LLMs strongly depends on the input data. Some methods exploit node-associated textual information and use the text-understanding ability of LLMs to improve graph representation learning \cite{guo2024graphedit}. Others assume a reasonably good pre-existing graph, similar to GNN-based methods, so that the LLM can reason over an already meaningful topology \cite{jin2024graph}.

\section{T2T-LA}

\subsection{Overview}

\begin{figure*}[!htbp] 
\centering
\includegraphics[scale=0.2]{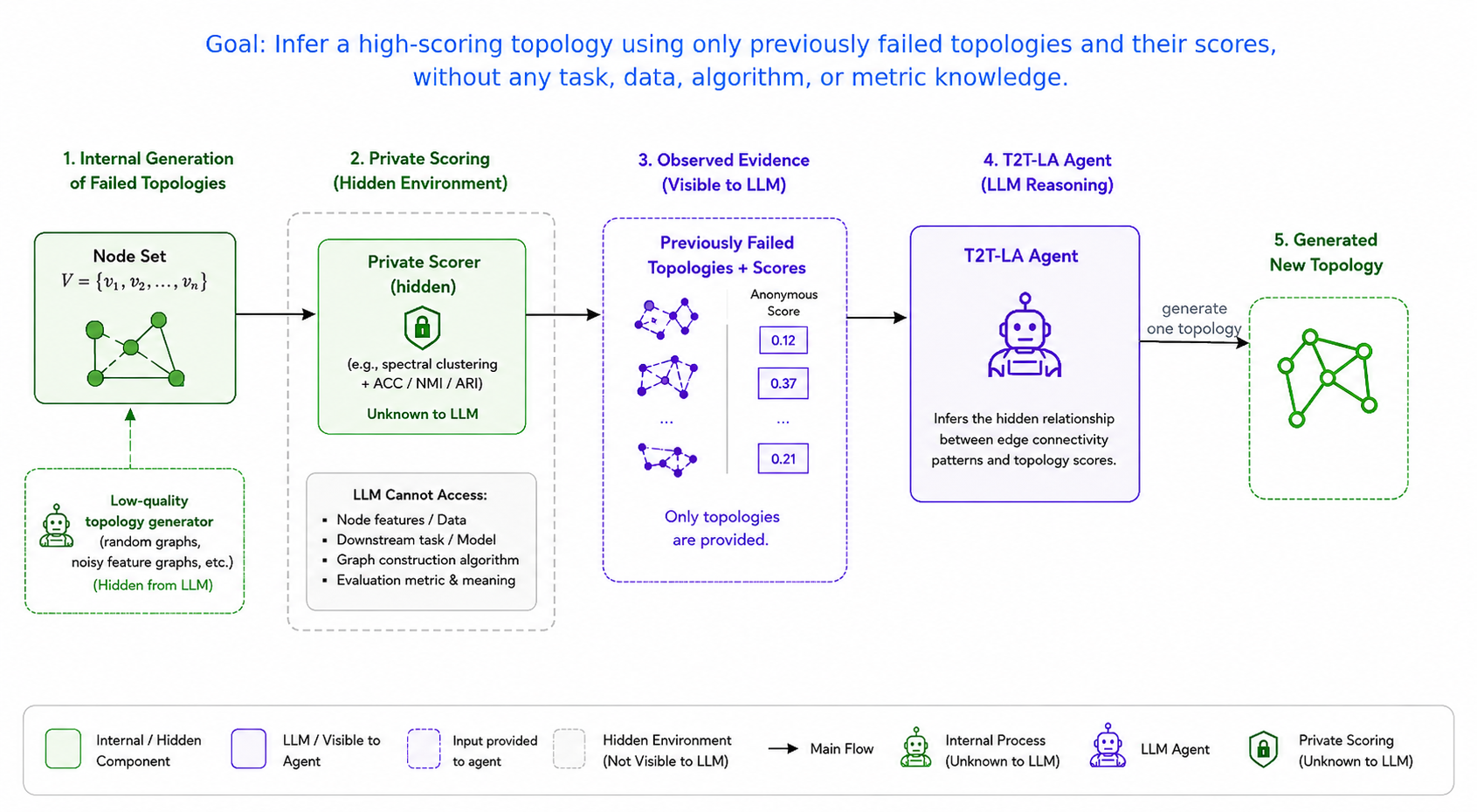}
\caption{Overview of T2T-LA. An internal low-quality topology generator first produces a set of failed graph topologies, which are evaluated by a private scorer. The T2T-LA agent receives only the previously failed topologies and their anonymous scores, without access to node features, downstream task information, graph construction rules, or metric meanings. Based on these topology-score pairs, the agent generates a new topology in a single pass.}
\label{fig:LLMGraphFlow}
\end{figure*}

Different from existing methods that apply LLMs to graph learning, our work does not use the understanding and reasoning abilities of LLMs to interpret sample feature vectors, such as textual node attributes, in order to obtain a better solution to the target problem. Instead, we use LLMs to analyze previously generated solutions and infer useful patterns from their structural differences. Therefore, our setting shifts the role of LLMs from data-level understanding to solution-level analysis and reasoning.

In this paper, we introduce T2T-LA, a Topology-to-Topology LLM Agent that reformulates graph topology learning as a solution-level reasoning problem. Figure~\ref{fig:LLMGraphFlow} illustrates the overall workflow of T2T-LA. The goal of the framework is to generate a graph topology that can help an unknown downstream algorithm achieve better performance, using only a set of failed topologies and their anonymous scores. Unlike conventional graph learning methods, T2T-LA does not receive the feature matrix, the downstream task, the downstream model, the graph construction rule, or the meaning of the evaluation metric. Instead, the agent is only provided with failed graph topologies and the anonymous scores assigned to them by a private scoring environment.

The workflow begins with an internal low-quality topology generator. Given a node set $V$, this generator produces a set of low-quality graph topologies using simple or noisy construction procedures. These procedures are not exposed to the LLM. Each generated topology is then evaluated by a private scorer. The private scorer assigns a score to each topology by running the target algorithm with that topology and evaluating the output using a predefined metric. The target task, the algorithm, and the metric are not revealed to T2T-LA.

It should be noted that, because these topologies are all generated by simple or noisy construction procedures, they can be reasonably expected to fail on the downstream task. However, if we find that a graph obtains a good score, we remove it from the input set.

After this hidden preparation stage, the visible input to T2T-LA consists only of previously failed topologies and their corresponding scores:
\[
    \{(A_1, s_1), (A_2, s_2), \ldots, (A_m, s_m)\}.
\]
Here, each $A_i$ denotes a graph topology and $s_i$ denotes its anonymous score assigned by the private scorer. The T2T-LA agent is then invoked once to generate a new topology $A^*$.

\subsection{Low-Quality Topology Generation}
This stage specifies how the failed input topologies are generated. In our implementation, we use two weak topology-generation procedures: random connected graph generation and noisy feature-local graph generation. The first procedure creates connected random graphs by first forming a random spanning structure and then adding random edges until a sampled average degree is reached. The sparsity level, degree heterogeneity, and triangle-closure ratio are randomly varied to produce diverse graph structures. The second procedure creates noisy feature-local graphs. It first constructs a nearest-neighbor graph in a PCA-projected feature space, and then injects noise by randomly dropping local edges, replacing some local edges with arbitrary random edges, and adding random long-range edges. A connectivity repair step is applied when necessary to ensure that the resulting graph is connected. These graphs are not optimized for the downstream task and are used only as topology-score observations for T2T-LA.

There are two reasons for using low-quality topologies as the input context for the LLM's reasoning process. First, this design prevents the agent from simply imitating high-quality examples. If a high-quality topology is provided, the LLM may copy or locally modify it without genuinely reasoning about the hidden relationship between graph topology and its score under an unknown downstream task. In contrast, by providing only failed topologies, T2T-LA is forced to reason from insufficient solutions. Although all observed topologies are poor and do not satisfy the desired quality requirement, their scores are not identical. This score variation among failed topologies provides useful signals. The agent must compare these poor topologies, identify structural differences that may explain their different scores, and infer potentially valuable connectivity patterns from them. Second, low-quality topologies are inexpensive to generate. Unlike carefully optimized graph construction methods, the random connected graph and noisy feature-local graph procedures used in our implementation do not involve explicit graph-learning objectives, iterative optimization, or task-specific parameter tuning. The former generates connected graphs through random edge construction, while the latter builds a local nearest-neighbor graph in a PCA-projected feature space and then injects random perturbations by deleting, replacing, and adding edges. Therefore, these procedures require little design effort and provide a cheap source of graph topologies for constructing topology-score observations for T2T-LA.

This stage is used only to create the input examples for T2T-LA. The generation process itself is hidden from the LLM. In particular, the LLM is not told whether the observed topologies are generated randomly, constructed from noisy feature relationships, or produced by any specific graph construction rule. Therefore, T2T-LA cannot rely on the topology-generation mechanism as prior knowledge. 


Formally, this stage produces a set of graph topologies
\[
    \mathcal{A}=\{A_1,A_2,\ldots,A_m\},
\]
where each $A_i \in \mathbb{R}^{n \times n}$ denotes an adjacency matrix over the node set $V$. The topologies in $\mathcal{A}$ are then passed to a private scorer for evaluation.

\subsection{Private Scoring}

After the low-quality topologies are generated, each topology is evaluated by a private scorer. The private scorer assigns a score to a topology by running the target algorithm with this topology on the target task and measuring the output using a predefined evaluation metric. Therefore, the score reflects the downstream performance obtained when the target algorithm uses the given graph topology.

For each topology $A_i \in \mathcal{A}$, the private scorer returns an anonymous scalar score:
\[
    s_i = \mathcal{S}(A_i),
\]
where $\mathcal{S}(\cdot)$ denotes the private scoring function. The target task, the target algorithm, and the evaluation metric are fixed inside the scorer, but they are not revealed to T2T-LA. As a result, T2T-LA only observes the topology-score pairs $(A_i,s_i)$ and treats each score as an anonymous indicator of the corresponding topology quality.

\subsection{Topology-to-Topology Reasoning Agent}

T2T-LA uses an off-the-shelf LLM as a topology-level reasoning agent. The LLM parameters are kept fixed, and no task-specific training or fine-tuning is performed. The visible topology-score pairs defined in the overview are provided only as in-context information.

The reasoning process is topology-to-topology: T2T-LA analyzes the observed structural patterns and their score variations, and then produces the new topology $A^*$ in a single pass. We refer to this setting as one-shot topology generation.

The one-shot design is central to T2T-LA. The agent is not expected to gradually improve a topology through trial-and-error feedback. Instead, because none of the observed graphs is sufficient by itself, the agent must compare the failed topologies and infer the structural regularities that explain why some of them are less poor than others. The purpose of this design is to make the agent form a topology-score understanding before producing the output, rather than relying on iterative correction after generation.

It should be emphasized that the objective of $A^*$ is not simply to obtain a higher score than the observed failed topologies. Those topologies are reasoning materials, not the final benchmark. The mission of T2T-LA is to generate a topology that is sufficiently useful for the hidden downstream graph-based task.

\subsection{Prompt Design}

The one-shot topology generation described above is implemented through a topology-only prompt. This prompt serves as the interface between the fixed LLM and the observed topology-score pairs. Rather than providing task-level instructions or data features, it specifies what information the agent is allowed to use and what form the generated output must take.

Fig.~\ref{fig:t2t_prompt} shows the core prompt used by T2T-LA. The prompt restricts the agent to graph topologies, topology descriptors, and anonymous scores. Here, topology descriptors refer to simple graph-level structural summaries, such as density, degree statistics, and connectivity information. The prompt also requires the output to be a Python function that constructs a new unweighted adjacency matrix.

\begin{figure}[!htbp]
\centering
\begin{tcolorbox}[
    colback=orange!12,
    colframe=black,
    width=0.98\columnwidth,
    boxrule=0.7pt,
    arc=0pt,
    left=3pt,
    right=3pt,
    top=3pt,
    bottom=3pt
]
\scriptsize
\ttfamily
Return Python code only. Define generate\_graph(n, observed\_graphs, observed\_profiles, helpers).

\vspace{1mm}
\normalfont\bfseries Role:\normalfont\ttfamily \\
You are T2T-LA, a topology-to-topology LLM agent.

\vspace{1mm}
\normalfont\bfseries Visible input:\normalfont\ttfamily \\
Anonymous graph topologies, topology descriptors, and anonymous scores.

\vspace{1mm}
\normalfont\bfseries Hidden information:\normalfont\ttfamily \\
No feature vectors, labels, graph construction methods, construction parameters, downstream task, downstream algorithm, or score semantics are provided.

\vspace{1mm}
\normalfont\bfseries Task:\normalfont\ttfamily \\
Compare failed topologies with different anonymous scores. Treat the score differences as reasoning clues, infer useful topology-level patterns, and generate a new unweighted graph topology for the hidden downstream task.
\vspace{1mm}
\normalfont\bfseries Constraints:\normalfont\ttfamily \\
Do not use features, labels, class membership, or any hidden task information. Do not simply return or lightly perturb an observed graph.

\vspace{1mm}
\normalfont\bfseries Required output:\normalfont\ttfamily \\
Return exactly one function generate\_graph(...), which outputs a symmetric, zero-diagonal, unweighted sparse adjacency matrix.
\end{tcolorbox}
\caption{Core T2T-LA prompt for topology-only one-shot graph generation.}
\label{fig:t2t_prompt}
\end{figure}

Fig.~\ref{fig:t2t_response} shows the corresponding response structure. The generated function ranks the observed failed topologies by their anonymous scores, contrasts structural patterns from lower-score and relatively higher-score failed graphs, and synthesizes a new topology rather than copying an existing one.

\begin{figure}[!htbp]
\centering
\begin{tcolorbox}[
    colback=green!14,
    colframe=black,
    width=0.98\columnwidth,
    boxrule=0.7pt,
    arc=0pt,
    left=3pt,
    right=3pt,
    top=3pt,
    bottom=3pt
]
\scriptsize
\ttfamily
def generate\_graph(n, observed\_graphs, observed\_profiles, helpers):\\
\hspace*{3mm}\# Topology-only blind-score reasoning.\\
\hspace*{3mm}\# 1. Rank observed graphs by anonymous scores.\\
\hspace*{3mm}\# 2. Treat higher-score graphs as positive structural evidence.\\
\hspace*{3mm}\# 3. Treat lower-score graphs as negative structural evidence.\\
\hspace*{3mm}\# 4. Score edges by contrastive support.\\
\hspace*{3mm}\# 5. Infer a sparse and balanced target degree.\\
\hspace*{3mm}\# 6. Construct a new graph instead of copying one.\\

\vspace{1mm}
\hspace*{3mm}profiles = sort\_by\_anonymous\_scores(observed\_profiles)\\
\hspace*{3mm}high\_profiles = profiles[:group\_size]\\
\hspace*{3mm}low\_profiles = profiles[-group\_size:]\\

\vspace{1mm}
\hspace*{3mm}edge\_score = contrastive\_edge\_support(\\
\hspace*{7mm}high\_profiles, low\_profiles, observed\_graphs)\\

\vspace{1mm}
\hspace*{3mm}selected = select\_positive\_edges\_with\_degree\_balance(edge\_score)\\
\hspace*{3mm}W\_new = build\_symmetric\_unweighted\_graph(selected)\\
\hspace*{3mm}W\_new = ensure\_connected\_random(W\_new, rng)\\
\hspace*{3mm}return W\_new
\end{tcolorbox}
\caption{Core LLM response structure. The generated function constructs a new topology by contrasting structural patterns between lower-score and higher-score observed graphs.}
\label{fig:t2t_response}
\end{figure}

\section{Evaluation}

\subsection{Experimental Setup}

\subsubsection{Downstream Task and Algorithm}

In this work, we use clustering as the downstream task. Clustering is an important problem in CAD and electronic system design, where complex design objects often need to be grouped according to hidden structural or behavioral similarities \cite{Singh2002EfficientCC}. For example, in cluster-based FPGA architectures, clustering is used to group several fine-grained logic blocks into a coarse-grained logic block. This organization improves FPGA utilization through dense local interconnects, while also shaping downstream testing and diagnosis problems because faults may involve both intra-cluster and extra-cluster connections \cite{Harris2000DiagnosisOI,Harris2000InterconnectTI}.

As the specific downstream algorithm for evaluating T2T-LA, we adopt Spectral Clustering (SC). SC is a powerful clustering algorithm that can discover complex non-convex structures and is highly useful in VLSI design \cite{Sajadinia2025HyperEF2S,Aghdaei2021HyperSFSH,Aghdaei2022HyperEFSH}.

\subsubsection{Evaluation Metrics}

Clustering Accuracy (ACC) measures the proportion of samples whose predicted cluster labels can be matched to the corresponding ground-truth classes. A higher ACC value indicates better clustering quality. Since cluster indices are arbitrary, the predicted labels must first be aligned with the true labels through an optimal permutation. Formally, ACC is defined as
\begin{equation}
\label{eq:acc}
ACC = \frac{1}{n}\sum_{i=1}^{n} 
\delta\!\left( y_i,\, \pi(c_i) \right),
\end{equation}
where $n$ is the number of samples, $y_i$ denotes the ground-truth label of sample $i$, and $c_i$ denotes its predicted cluster assignment. The indicator function $\delta(x,y)$ equals $1$ if $x=y$ and $0$ otherwise. The mapping $\pi(\cdot)$ is the optimal permutation between cluster indices and class labels, which is typically obtained using the Hungarian algorithm~\cite{premachandran2013consensus}.

\subsubsection{Low-quality Graphs}
For each dataset, we build a fixed pool of 380 low-quality graph topologies, including 120 random connected graphs and 260 noisy feature-local graphs. Random connected graphs provide feature-independent topology diversity, while noisy feature-local graphs provide perturbed local structures loosely related to feature neighborhoods. We use more noisy feature-local graphs to cover a broader range of degraded local patterns, and use random connected graphs to diversify the pool beyond feature-local constructions.

To reduce the computational and prompting burden of the agent, we do not expose the entire pool of 380 topologies to T2T-LA. Instead, after evaluating the internally generated topologies with the downstream spectral clustering algorithm, we select a compact ACC-stratified subset as the observed topology set. Specifically, we divide the low-to-medium accuracy range into five intervals, $[10\%,20\%)$, $[20\%,30\%)$, $[30\%,40\%)$, $[40\%,50\%)$, and $[50\%,60\%)$. Within each interval, we sort all available topologies by ACC in descending order and select eight evenly spaced topologies from the ranked list. As a result, 40 observed topologies are provided to the agent for each dataset. Topologies with ACC above $60\%$ are not exposed to the agent.

\subsubsection{LLM}

In our experiments, T2T-LA is implemented using GPT-5.5 Thinking as the LLM agent. 

\subsection{Datasets}

To evaluate the topology-generation mechanism of T2T-LA in a controlled and reproducible setting, we use standard benchmark datasets that are widely adopted in spectral clustering studies. These datasets are carefully curated for algorithm evaluation: their sample selection, feature representations, and class labels are well defined. This allows the evaluation to focus on the behavior of the proposed method itself, rather than being distorted by domain-specific annotation noise, severe sample imbalance, or other dataset-specific uncertainties. Therefore, they provide a clean testbed for evaluating the topology-generation mechanism of T2T-LA without conflating algorithmic performance with dataset-specific biases.
We adopt the following three well-established benchmark datasets: PenDigits, USPS, and MNIST.
\begin{itemize}

    \item \textbf{PenDigits:} PenDigits contains $7{,}494$ samples from $10$ classes. Each sample is represented by a $16$-dimensional numerical feature vector, making it a compact benchmark for evaluating clustering performance on low-dimensional structured data.

    \item \textbf{USPS:} USPS contains $9{,}298$ samples from $10$ classes. Each sample is represented by a $256$-dimensional feature vector, providing a medium-dimensional benchmark for testing graph-based clustering methods.

    \item \textbf{MNIST:} MNIST contains $70{,}000$ samples from $10$ classes. Each sample is represented by a $784$-dimensional feature vector, making it a larger-scale benchmark for evaluating clustering algorithms on high-dimensional data.

\end{itemize}

\subsection{Compared Algorithms}

In our evaluation, we not only want to test whether the graph generated by T2T-LA can enable SC to produce a sufficiently good solution, but also want to examine how the graph generated by T2T-LA without access to data features or any information about the downstream task and algorithm compares with state-of-the-art graph construction methods that use data features and know the downstream task and algorithm. To this end, we evaluate T2T-LA against the following baseline and state-of-the-art methods.

\begin{itemize}

    \item \textbf{Standard $k$NN}: This baseline constructs an affinity graph by connecting each sample to its $k$ nearest neighbors in the original feature space. The resulting graph captures local neighborhood relationships and serves as a common graph construction method for spectral clustering.

    \item \textbf{Consensus $k$NN (cons-$k$NN)}~\cite{premachandran2013consensus}: This method improves the conventional $k$NN graph by identifying more reliable neighbor relationships. It uses neighborhood consensus information to select strong connections and remove less reliable edges, thereby producing a more robust affinity graph.

    \item \textbf{Spectral edge sparsification}~\cite{DBLP:conf/bmvc/WangF22}: This approach sparsifies a graph by removing redundant edges while preserving important spectral characteristics. The goal is to reduce graph complexity without significantly changing the spectral structure that is important for downstream graph-based algorithms.

    \item \textbf{LSGL}~\cite{DBLP:conf/iclr/KalofoliasP19}: LSGL is a graph-signal-processing-based graph learning method. It extends the graph learning model in~\cite{kalofolias2016learn} by automatically adjusting model parameters.

\end{itemize}

\subsection{Numerical Results}

Table~\ref{table:compareACC} reports the clustering accuracy of different graph construction methods on COIL20, PenDigits, and USPS. From the results, we can observe that T2T-LA is able to generate highly effective graph topologies by reasoning from only 40 weak topologies whose ACC values are all below 60\%. Across COIL20, PenDigits, and USPS, the graphs generated by T2T-LA allow spectral clustering to achieve ACC values that are more than 25, 27, and nearly 17 percentage points higher than the maximum ACC of the topologies used as reasoning materials, respectively. These results demonstrate that T2T-LA can effectively infer useful hidden structural patterns from weak and previously failed topologies, showing the effectiveness and promise of the proposed framework.

Compared with SOTA graph construction and graph learning methods, T2T-LA also demonstrates strong superiority. Among the three datasets, T2T-LA outperforms all baseline and SOTA methods on two datasets, namely COIL20 and PenDigits. On the remaining dataset, USPS, T2T-LA does not achieve the best performance, but still ranks second. Notably, its accuracy on USPS is less than 5 percentage points lower than LSGL, whereas on PenDigits, T2T-LA exceeds LSGL by more than 12 percentage points. These results indicate that, without using data feature vectors, without knowing the downstream task or algorithm, and without relying on any hand-crafted rules, T2T-LA can still outperform feature-dependent and theoretically sophisticated SOTA methods.
\begin{table}[t]
\centering
\caption{Clustering accuracy (\%)}
\label{table:compareACC}
\small
\renewcommand{\arraystretch}{1.15}
\setlength{\tabcolsep}{3pt}
\resizebox{\columnwidth}{!}{
\begin{tabular}{lccccc}
\hline
Dataset & $k$-NN & Consensus & Spectral Spar. & LSGL & T2T-LA \\
\hline
COIL20    & 75.72 & 81.60 & 76.27 & \underline{85.49} & \textbf{85.76} \\
PenDigits & 74.36 & 71.08 & \underline{83.26} & 74.53 & \textbf{87.07} \\
USPS      & 64.31 & 68.54 & 70.74 & \textbf{81.50} & \underline{76.67} \\
\hline
\end{tabular}
}
\end{table}

\section{Conclusion}
We introduced T2T-LA, an agentic framework that leverages the strong reasoning capability of LLMs to extract useful structural information from previously failed cases and generate high-quality graph topologies for graph algorithms. T2T-LA does not rely on data features or hand-designed algorithmic rules, nor does it require prior knowledge of the downstream algorithm. Instead, it only needs a small number of previously failed topologies and their corresponding failure measurements to complete the topology generation task. Experiments on COIL20, PenDigits, and USPS demonstrate that T2T-LA can generate highly competitive graph topologies and outperform SOTA methods that rely on data feature matrices, sophisticated mathematical theory, and carefully designed graph learning algorithms.

\end{document}